\documentclass[11pt,a4paper]{article}
\usepackage[T1]{fontenc}
\usepackage{lmodern,microtype}
\usepackage[margin=26mm]{geometry}
\usepackage{amsmath,amssymb,amsthm,booktabs,tabularx,array,enumitem}
\usepackage{xcolor,graphicx,placeins}
\usepackage[numbers,sort&compress]{natbib}
\usepackage{lineno}
\usepackage{hyperref}
\definecolor{ink}{HTML}{203946}
\definecolor{linkteal}{HTML}{235C63}
\hypersetup{colorlinks=true,linkcolor=linkteal,citecolor=linkteal,urlcolor=linkteal,
pdftitle={puma.ai: Poomsae Uncertainty-Aware Movement Alignment},
pdfauthor={Keivan Shariatmadar, Ahmad Osman, Kambiz R. Ali}}
\newcommand{\doi}[1]{doi: \href{https://doi.org/#1}{\nolinkurl{#1}}}
\setlist{leftmargin=*,itemsep=2pt}
\newtheorem{proposition}{Proposition}
\newcommand{\puma}{puma\ensuremath{\cdot}ai}
\newcommand{\NA}{\bot}
\newcommand{\ind}{\mathbf{1}}
\newcommand{\R}{\mathbb{R}}
\newcommand{\E}{\mathcal{E}}
\newcolumntype{Y}{>{\raggedright\arraybackslash}X}

\title{\Large \puma: Poomsae Uncertainty-Aware Movement Alignment\\[5pt]
\normalsize An Evidence Framework for Recognized Taekwondo Poomsae}
\author{Keivan Shariatmadar\textsuperscript{1}\thanks{Correspondence: \href{mailto:keivan.shariatmadar@gmail.com}{keivan.shariatmadar@gmail.com}. WT and ETU Education Committee Member and Technical Delegate. Luxembourg Referee Chair.}~~
Kambiz R. Ali MBE\textsuperscript{2}}
\date{\small
\textsuperscript{1}AutomaTiQ, htw saar University of Applied Sciences, Saarbr\"ucken, Germany\\
\textsuperscript{2}British Taekwondo, Poomsae Referee Chair, United Kingdom\\[4pt]
Hypothesis and Theory}
\begin{document}
\maketitle
\begin{abstract}
Recognized Taekwondo Poomsae consists of prescribed forms whose ordered movements provide a natural reference for video analysis. A pose trajectory, however, can be ambiguous because of a missed track, an obscured joint, a segmentation error, or an ambiguity in how a performance aligns to the reference. We propose Poomsae Uncertainty-Aware Movement Alignment (\puma), a shared video-analysis method with an expert-curated, versioned reference for each form. Movement-level edit alignment allows unmatched observed segments and unmatched reference occurrences. A set of near-optimal paths represents alternative alignments, and a separate visibility test determines whether the video supports a movement-level assessment. We derive an exact forward--backward test for individual edge support and bound its sensitivity to cost perturbations on a fixed edit graph. Exhaustive enumeration of an abstract example shows how a possible extra movement can be localized to an interval while its attribution to a particular segment remains ambiguous. The proposed evaluation starts with Taegeuk Il Jang and combines athlete-disjoint testing, a paired camera comparison, and a controlled reader study with qualified judges. The intended outputs are reviewable clips and auditable coaching or education prompts. The framework yields a testable hypothesis: at comparable decision coverage under matched observation conditions, retaining plausible alignments will reduce unsupported movement assertions relative to single-best alignment under the same evidence gate. The paper develops the formulation and proposed tests. Empirical performance and benefits to users remain to be measured.
\end{abstract}

\section{Introduction}
The same set of movements can be performed at different tempos, and a similar-looking video sequence can arise from different movement orders or camera views. Recognized Poomsae narrows the analysis problem because the intended form is known. Its reference sequence can constrain where the system looks and how it associates observed segments with expected movements. For a coach, a useful output is the span of video in which a transition warrants review. For a judge educator, it is a reproducible case with the relevant occurrence, camera view, and reasons for uncertainty.

World Taekwondo (WT) distinguishes accuracy from presentation in recognized Poomsae. Its published rulebook allocates four points to accuracy and six to presentation. Accuracy includes basic and form-specific movements and balance. Presentation includes speed and power, rhythm and tempo, and expression of energy \citep[Arts.~15--16]{wtRules}. The recognized Poomsae handbook provides examples of wrong, extra, and omitted techniques \citep[pp.~20--23]{wtHandbook}. This study considers whether a video analysis system can locate observable sequence deviations and provide evidence for qualified reviewers. Video may support kinematic observations, while application of a rule to a case and evaluation of presentation require sport-specific judgment. Estimated motion speed alone provides no direct measure of mechanical power or force.

We develop an occurrence-level alignment model that retains alternative explanations and an explicit option to withhold an assessment. We treat coaching and judge education as separate outcomes. A system can improve retrieval of the relevant movement without improving the judge's interpretation. The proposed experiments measure both. The primary hypothesis of \puma{} is that, at comparable decision coverage, retaining alternative alignments within an evidence-gated reviewer yields fewer unsupported assertions than using a single best path with the same pose inputs, reference examples and evidence gate. We distinguish an incorrect assertion on a reference-labeled case from a definite assertion made with inadequate video. The planned evaluation measures each separately. A paired camera experiment tests whether a second view improves coverage at comparable risk. A reader study tests whether the cues improve reference-supported reasoning.

\begin{figure}[tbp]
\centering
\includegraphics[width=\linewidth]{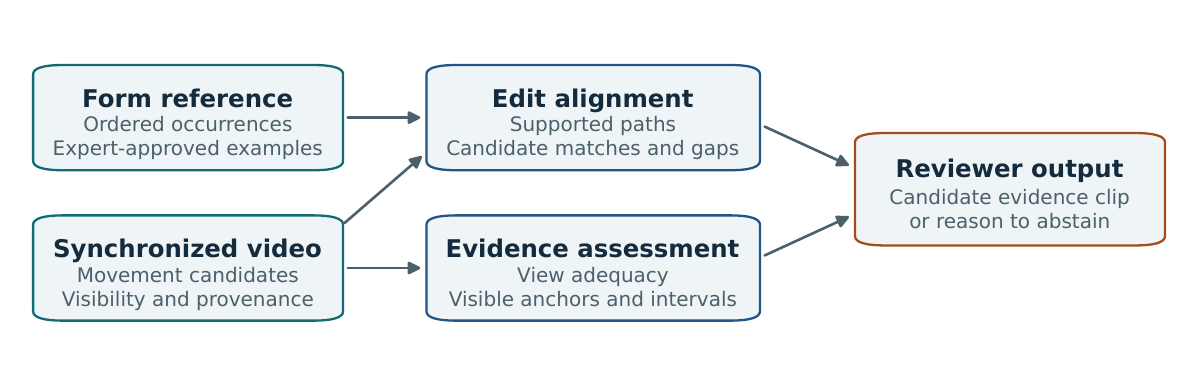}
\caption{\puma{} workflow. The form reference constrains alignment. The evidence assessment separately checks whether the available video supports the observation that would appear in the review record.}
\label{fig:workflow}
\end{figure}

\section{Related work and positioning}
\citet{hoang2023} use estimated three-dimensional poses and dynamic time warping (DTW) to align performances of Taegeuk Il Jang with a reference. Their separation of posture from body proportions is directly relevant to feature design. Their approach motivates the single-reference DTW baseline. DTW aligns time series under local path constraints \citep{sakoe1978}. In the proposed analysis, explicit gap transitions make potentially unmatched occurrences and observations available for review.

Related work in diving provides a useful precedent for evaluating correspondence at the level of procedural steps. FineDiving includes annotations of consecutive steps and aligns steps between performances for quality assessment \citep{xu2022}. Its annotations and scoring target are specific to diving. Here the movement occurrence is the unit of alignment, and the initial output is an observation with supporting evidence. Temporal action segmentation addresses a related problem: competing partitions can split a single action or merge two actions. Learned methods such as MS-TCN offer a future candidate-generation comparison once sufficient in-domain annotations are available \citep{abufarha2019}. The first experiment uses a fixed segmentation baseline and a human-boundary comparison.

\citet{lee2020} report an RGB-D dataset of individual Taekwondo techniques. Unit-technique clips can support component-level analysis, while alignment and omission detection require ordered routines and occurrence-level annotations. \citet{huang2026} release synchronized frontal and lateral recordings for Taegeuk 1--8, accompanied by frame-aligned two-dimensional pose estimates. These data can support ingestion, synchronized-view workflow checks and exploratory alignment, within the academic and educational use described by the article, with permission and use terms to be confirmed before reuse. The data article describes controlled capture and estimates the landmarks with MediaPipe. The released joints alone cannot furnish an adjudicated record of rule-relevant omissions or supply calibrated three-dimensional reference measurements.

\citet{choi2025} evaluate neural networks for classifying labeled Poomsae postures. Frame classification supports recognizing posture categories, while omission and repetition depend on a segment's position in the full form. In a case study of Pyongwon, \citet{arriola2025} compare computer vision-based accuracy evaluation with judges' assessments. \citet{kim2026} analyze recognized Poomsae score sheets by scoring item and competition stage. Such score summaries support analyzing scoring criteria separately, while providing limited information about the correctness or visibility of a particular movement. \citet{jeon2026} examine judging reliability in freestyle Poomsae. Their work motivates distinguishing agreement and repeatability in study design, with neither their task nor their reliability estimates assumed to transfer to recognized Poomsae. A systematic review by \citet{hua2026} distinguishes recognition and movement measurement from the evidence required to support scores or rule-based deductions. Taken together, prior work motivates a reference comparison and a separate, movement-level assessment of what the system has localized and what the video can substantiate. The proposed research tests those two capabilities directly.

\section{Scope and intended applications}
The initial setting is an offline review of individual recognized Poomsae performed by consenting adults. The form identifier is supplied by the coach or session record. The first research form is Taegeuk Il Jang. A second form such as Koryo is a follow-on study requiring a new expert reference and a new validation set. The proposal preserves a distinction between movement observation, educational feedback and official scoring. Manual reviewers retain decision authority, and the initial system has no scoring or results write access.

\begin{table}[htbp]
\centering\small
\caption{Potential uses and the level of evidence required.}
\label{tab:uses}
\begin{tabularx}{\textwidth}{@{}>{\raggedright\arraybackslash}p{.25\textwidth}YY@{}}
\toprule
Use & Proposed output & Assessment scope\\
\midrule
Sequence practice & Clip of a possible repeated, omitted, or unexpected segment & Initial pilot focuses on extra and omitted segments.\\
Coaching movement review & Aligned clips for a visible stance, transition, or direction change & Coach verifies the observation and selects feedback.\\
Judge education & Shared cases with original video and independent reviewer rationales & Initial reader study tests the effect of AI suggestions.\\
Pair and team practice & Offset between corresponding movement onsets & Requires per-person tracking and separate validation.\\
Athlete progress & Trends in coach-confirmed observations under comparable capture conditions & Requires repeat-session data.\\
Event video management & Read-only indexing by performance ID, form, and court & Requires authorized exports and event reconciliation.\\
\bottomrule
\end{tabularx}
\end{table}

Freestyle has a different technical and presentation framework \citep[Arts.~9 and 15--16]{wtRules}. Para Poomsae has a separate rulebook \citep{wtPara}. A future Para study would require co-design with Para athletes and relevant officials, with references reflecting the applicable category and eligible adaptations. The present formulation does not infer a sport class. If an event adopts a specific interpretation or rule edition, reviewers must record that edition in the study protocol and reference manifest.

\section{Form references and observed movement segments}
Let \(f\) be a recognized form and \(r\) the reference edition, which records the applicable rulebook and interpretation. Define
\[
\mathcal{T}_{f,r}=(q_1,\ldots,q_J),\qquad
q_j=(j,\kappa_j,\sigma_j,\ell_j,\theta_j,\E_j).
\]
Here \(\kappa_j\) denotes the technique or group of simultaneous actions, \(\sigma_j\) the stance, \(\ell_j\) laterality, \(\theta_j\) the expected direction and \(\E_j\) a set of expert-accepted examples. The index \(j\) identifies an occurrence even if the same technique appears at multiple positions. The reference records sources, training examples and permissible variation agreed by reviewers. Explicitly marked disagreements stay in the annotation record. If reviewers accept alternate sequence structures, each approved structure can be evaluated on its own ordered graph. The initial pilot uses a fixed sequence.

A shared pose estimator yields time-stamped image landmarks. Candidate movement intervals \(I_i=[s_i,e_i]\), \(i=1,\dots,N\), are ordered and non-overlapping in the initial formulation. For a reproducible baseline, smooth visible joint-speed traces within contiguous tracked intervals, propose boundaries at local minima of a pooled motion-energy trace with minimum spacing, and retain the intervening intervals as candidates. Choose smoothing, spacing and prominence on development data. Sustained low motion may be an intentional hold, and missed tracks break the trace rather than count as low speed. Test the resulting segmentation against expert boundaries and keep a human-boundary comparison to identify propagation of segmentation errors. We distinguish apparent body orientation and laterality from motion in camera coordinates and retain original timestamps for timing analyses. Per-view descriptors may include observable joint angles and direction changes where justified by the view. Encode cyclic orientations with sine and cosine components or use a wrapped angular residual so that crossing the angle origin does not inflate the distance. A two-dimensional projection generally does not identify a unique three-dimensional pose. In a Taegeuk Il Jang study, \citet{fukushima2024} compared joint angles from calibrated multiview OpenPose triangulation with VICON and found that errors differed across joint regions. This motivates verifying each cue's measurement performance under the planned acquisition setup. View fusion therefore operates on synchronized evidence and comparable features, with 3D reconstruction requiring a separately calibrated acquisition system.

Let \(x_i\in\R^d\) describe segment \(i\), and \(O_i\subseteq\{1,\ldots,d\}\) denote its observable feature indices. Pose-model scores can help build a mask but require validation against manual visibility annotations. They are not probabilities that a technique was correctly executed. Write \(z_{jk}\) for expert exemplar \(k\) of occurrence \(j\), with available coordinates \(O_{jk}^{\rm ref}\). The intersection is \(A_{ijk}=O_i\cap O_{jk}^{\rm ref}\). Use nonnegative weights \(w_m\), strictly positive scales \(a_m\), and a bounded loss \(\rho(u)=\min(|u|,c)\) with \(c>0\). For exemplar pairs whose overlap meets a prespecified minimum and contains the features needed for comparison, define
\begin{equation}
d_{ij}=\min_{k\in K_{ij}}
 \frac{\sum_{m\in A_{ijk}} w_m\,
 \rho((x_{im}-z_{jkm})/a_m)}
 {\sum_{m\in A_{ijk}}w_m}.
\label{eq:cost}
\end{equation}
The nonempty-overlap and weight conditions guarantee a positive denominator for every \(k\in K_{ij}\). If this set is empty, use a prespecified finite unknown-match cost and mark the entry as unsupported for assessment. It is useful to vary that cost in sensitivity analysis because a very low or high default can change the inferred edit path. Fit scales, weights and thresholds with development data only, and control exemplar counts or compare them explicitly because the minimum in Eq.~\eqref{eq:cost} tends to favor larger reference sets. The output is a dissimilarity, with no interpretation as a deduction or calibrated probability.

\subsection{Timing, visibility and comparable measurements}
For camera \(v\), let \(t_v\) be a local video timestamp and let \(t=a_vt_v+b_v\) be its mapping to the shared reference clock. A shared clock may make \(a_v=1\) appropriate. Otherwise, two or more synchronization events at different times allow drift to be checked. Define the alignment residual for check event \(h\) as \(r_{vh}=a_vt_{vh}+b_v-t_h^{\mathrm{ref}}\). Report absolute residuals and missing synchronization events. Withhold a two-view decision when synchronization cannot be verified within the preregistered tolerance for the movement criterion. Frame timestamps, synchronization residuals and original clip windows are retained for review.

Use view-specific features and reference examples tagged by capture setup. Normalize image-plane lengths by a prespecified scale measured from sufficiently tracked frames. Keep camera direction and athlete laterality in separate fields. Where a scale estimate or left--right identity is unreliable, drop only the affected features and record the reason. When comparing two views, calculate per-view distances on the same candidate interval. One proposed fusion policy averages the finite per-view costs over usable views with fixed, development-set weights, retaining which views contributed to each comparison. A task can require a particular view or cue even if the aligner can assign a finite cost from another view. This distinction is represented by the separate evidence gate in Fig.~\ref{fig:workflow}.

\section{Edit alignment and evidence sufficiency}
\subsection{Minimum-cost alignment}
Construct a directed acyclic graph from \(s=(0,0)\) to \(t=(N,J)\). The state \((i,j)\) denotes consumed observed segments and expected occurrences. A diagonal edge matches the next segment to the next occurrence. An observation-only edge leaves a segment unmatched (\emph{insertion}), and a reference-only edge leaves an occurrence unmatched (\emph{deletion}). With positive gap costs \(\lambda_I,\lambda_D\), the minimum forward cost is
\begin{equation}
F(i,j)=\min\left\{
\begin{array}{l}
F(i-1,j-1)+d_{ij},\\
F(i-1,j)+\lambda_I,\\
F(i,j-1)+\lambda_D
\end{array}
\right.,
\label{eq:dp}
\end{equation}
with \(F(0,0)=0\), \(F(i,0)=i\lambda_I\), \(F(0,j)=j\lambda_D\). The recurrence is a movement-level instance of global sequence alignment with gaps \citep{needleman1970}. A gap represents a relation in the model. An extra observation may be a split segment, and a reference gap may originate in a merged segment or lost track. An out-of-order pair may appear as an insertion and a deletion. The model therefore treats order changes as cases for review without asserting a dedicated transposition detector.

\subsection{Alternative paths and exact edge support}
For fixed features, template and gap costs, let \(C(\pi)\) be the sum of edge costs along path \(\pi\), and let
\begin{equation}
\Pi_\delta=\{\pi:s\leadsto t,\ C(\pi)\le F(t)+\delta\}, \qquad \delta\ge0 .
\label{eq:paths}
\end{equation}
This is a set of plausible alignments under a specified cost and tolerance. The tolerance is in cost units and is fitted with development data. It provides no nominal statistical coverage. Let \(B(v)\) be the shortest cost from state \(v\) to \(t\).
\begin{proposition}
An edge \(e=(u,v)\) is supported by at least one path in \(\Pi_\delta\) if and only if
\[
F(u)+c(e)+B(v)\le F(t)+\delta.
\]
\end{proposition}
\begin{proof}
Any path that visits \(e\) pays at least \(F(u)+c(e)+B(v)\). Conversely, concatenating shortest prefix and suffix paths with \(e\) attains the bound. Topological order of the edit graph makes this concatenation a path. The claimed condition is equivalent to its cost lying within the specified tolerance.
\end{proof}
A forward pass, a backward pass and one edge scan obtain the support of each edge in \(O(NJ)\) time and memory, excluding feature extraction and costs. For expected occurrence \(j\), let \(\mathcal{L}_j\) be the labels projected from the supported edges that consume it, with label \(i\) for a match and label \(\varnothing\) for an unmatched occurrence. A singleton \(\{i\}\) establishes stable correspondence relative to \(\Pi_\delta\). A singleton \(\{\varnothing\}\) establishes that the model always leaves the occurrence unmatched. Sets combining different observed indices indicate localization uncertainty, and sets combining an index and \(\varnothing\) indicate uncertainty about the existence of a match. Joint claims require a constrained path query because individually supported edges may occur on different paths. For a binary event predicate \(P\) implemented by a small state machine on the edit graph, it holds on every admissible path if the minimum cost of any path violating \(P\) exceeds \(F(t)+\delta\).

When multiple views or expert-vetted variants are plausible, \(\Pi_\delta\) can be computed separately for each explicitly defined scenario. An assessment can then require the same conclusion across scenarios judged applicable, with the tolerance interpreted relative to each scenario's optimum. The primary experiment assesses a fixed policy and uses a prespecified sensitivity analysis to vary the mask, unknown cost, segmentation and gap costs. An omitted source or unavailable view remains visible in the audit record.

\subsection{Sensitivity to tolerance and cost uncertainty}
Define the slack of an edge as
\begin{equation}
m(e)=F(u)+c(e)+B(v)-F(t).
\label{eq:slack}
\end{equation}
The slack is nonnegative and gives the minimum tolerance required to admit the edge. If \(\delta_1\le\delta_2\), then \(\Pi_{\delta_1}\subseteq\Pi_{\delta_2}\). With the gate and the map from paths to labels held fixed, a singleton label set can grow as tolerance increases, and a larger set cannot become a singleton. Coverage therefore cannot increase as a function of \(\delta\) on fixed opportunities under this policy. The empirical error rate among issued assessments can increase or decrease because the cases removed may differ in difficulty.

\begin{proposition}
Suppose the graph is fixed and the cost of each edge changes by at most \(\varepsilon\). For any edge \(e\), the perturbed slack \(m'(e)\) satisfies
\[
|m'(e)-m(e)|\le 2(N+J)\varepsilon .
\]
\end{proposition}
\begin{proof}
Every edge consumes at least one index, so any complete path has at most \(L=N+J\) edges. Its cost changes by at most \(L\varepsilon\). Both the minimum over all paths and the minimum over paths through \(e\) can therefore change by at most \(L\varepsilon\), since the sets of paths are fixed. Subtracting the two minima gives the bound.
\end{proof}
For fixed \(\delta\), an edge with \(m(e)<\delta-2L\varepsilon\) is guaranteed to remain supported. An edge with \(m(e)>\delta+2L\varepsilon\) is guaranteed to remain excluded. Edges between these two limits require further analysis. The inequality bounds sensitivity to a specified cost perturbation. It does not quantify the probability of error or cover changes to the graph caused by candidate segmentation, and it does not replace a visibility check when a camera becomes unavailable.

\subsection{Selective assessment}
For a task-specific opportunity \(u\), evaluate adequate visual evidence \(G(u)\in\{0,1\}\), and let \(H(u)\) be the set of propositions projected from plausible paths for that opportunity, using a predefined label vocabulary \(\mathcal{Y}_u\). Define
\begin{equation}
\widehat{y}(u)=
\begin{cases}
y,&G(u)=1\text{ and }H(u)=\{y\},\quad y\in\mathcal{Y}_u,\\
\NA,&\text{otherwise}.
\end{cases}
\label{eq:reject}
\end{equation}
A reference gap can yield a possible omission only when its interval and bracketing movements are identifiable, the relevant motion is observable through the interval, and tracking or segmentation failure is not a plausible explanation under the prespecified gate. Near the start or end of a routine, use a verified recording boundary as an anchor only if the footage includes the complete relevant interval. A possible extra segment also requires corroborating sequence context. An abstention records whether the problem is visibility, missing anchors, segmentation or alignment ambiguity, or an unavailable input. This use of rejection motivates evaluation with risk and coverage together \citep{geifman2017}. It transfers no guarantee from that work.

\subsection{Symbolic analysis of a repeated segment}\label{sec:example}
Consider the synthetic reference \(A,B,C\) and observation \(A,B_1,B_2,C\), where \(B_1\) and \(B_2\) have the same movement descriptor but distinct timestamps. Let identical labels cost zero, other matches cost three, and each gap cost one. With \(\delta=0\), the optimum cost is one. The observations can be reconciled by inserting either \(B_1\) or \(B_2\), and both routes are optimal. The two observed segments can be shown together for review if the video supports the interval. The alignment cannot establish which \(B\) to label as extra. The support test returns both insertion locations. A constrained query verifies that every minimum-cost path inserts exactly one segment within that pair. A visually obscured turn is a different failure: even agreement among optimal paths cannot make its stance assessable. These examples illustrate the decision logic without estimating performance on athletes or videos.

\begin{figure}[!tbp]
\centering
\includegraphics[width=.84\linewidth]{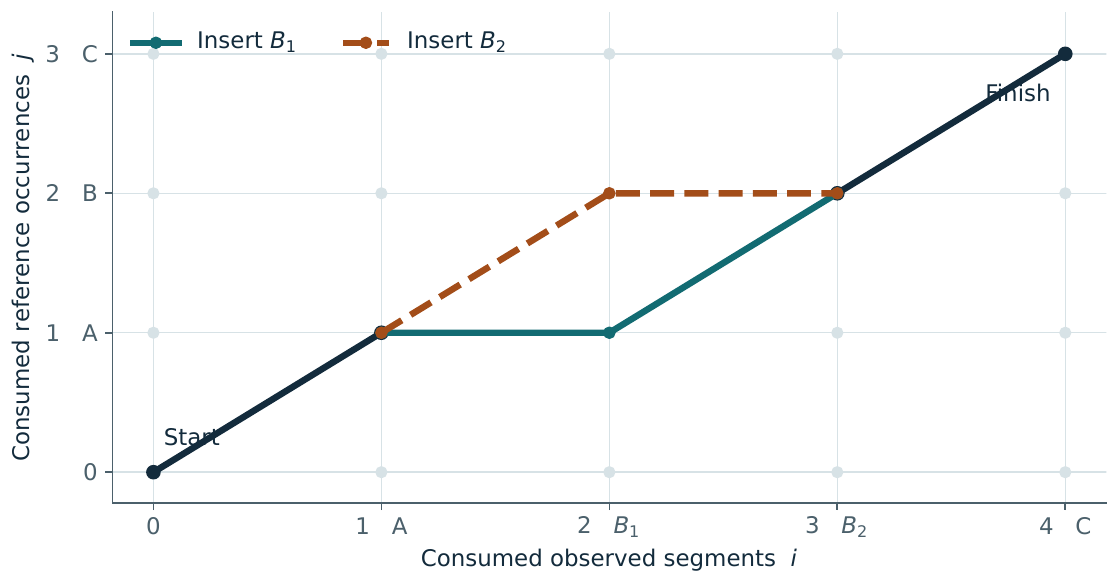}
\caption{Two minimum-cost paths in the symbolic example. Horizontal edges consume an observed segment without matching it to a reference occurrence. A diagonal edge consumes one of each. The paths agree about the affected pair but identify different segments as unmatched. Only the optimal edges are highlighted.}
\label{fig:paths}
\end{figure}

\begin{figure}[!htbp]
\centering
\includegraphics[width=.95\linewidth]{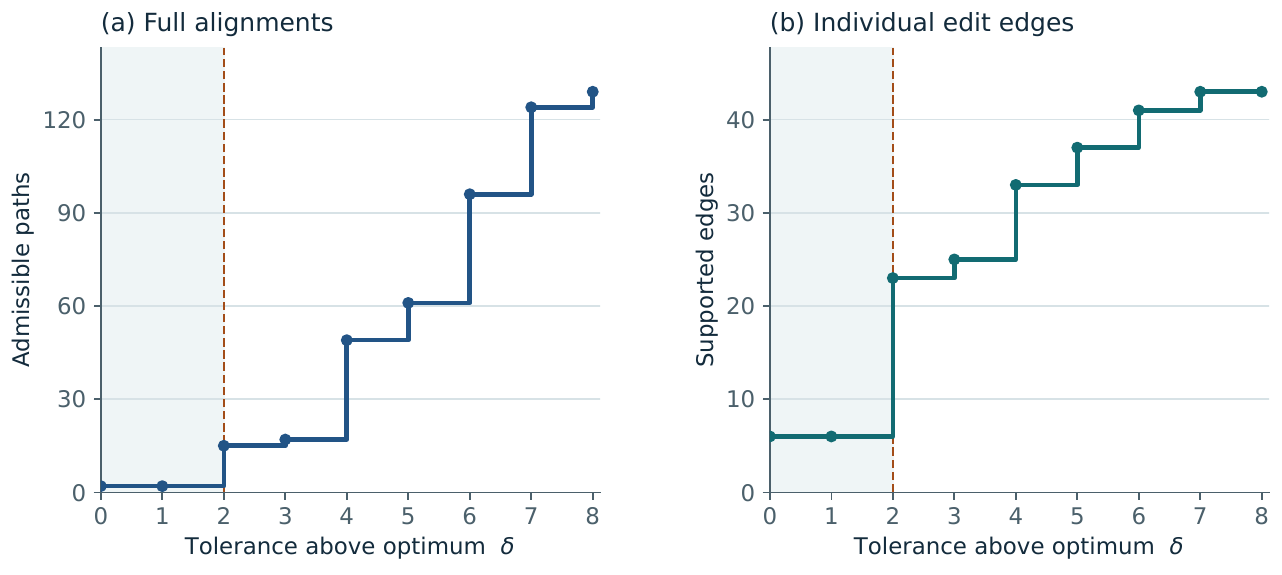}
\caption{Exact tolerance sweep for the symbolic example. The dashed line marks the first tolerance admitting an alignment with a deletion. To its left, every admissible path leaves one segment in the repeated pair unmatched. The two panels distinguish whole-path ambiguity from the number of edges that can appear in at least one such path.}
\label{fig:sweep}
\end{figure}

The localization ambiguity in Fig.~\ref{fig:paths} persists for all nonnegative tolerances. A stronger conclusion, that every admissible path leaves exactly one of the two \(B\) segments unmatched and has no deletion, holds only within a specific tolerance range. To see this, let a mismatched diagonal have cost \(\mu>0\), and write \(\lambda_I,\lambda_D>0\) for the two gap costs. If there are \(d\) deletions, the path must have \(d+1\) insertions because the observation has one more segment than the reference. Hence \(F(t)=\lambda_I\) and a path with a deletion has excess cost at least \(\lambda_I+\lambda_D\). This bound is attained by replacing one zero-cost diagonal in an optimal path with a deletion and an insertion. A deletion-free path can insert outside the \(B\) pair only by leaving \(BBC\) or \(ABB\) to match against \(ABC\). Either case requires exactly one mismatch, at an excess cost of \(\mu\). Consequently, the stronger conclusion holds when
\begin{equation}
0\le \delta < \min\{\mu,\lambda_I+\lambda_D\}.
\label{eq:toy-threshold}
\end{equation}
Equality admits a counterexample. In the plotted example \((\mu,\lambda_I,\lambda_D)=(3,1,1)\). At \(\delta=0\) there are two admissible paths using six distinct edges. At \(\delta=2\) there are 15 admissible paths using 23 distinct edges. Figure~\ref{fig:sweep} reports the complete sweep over the distinct integer tolerances. The fully admitted graph contains 129 paths. These counts describe the symbolic graph only.

\FloatBarrier

\section{Proposed empirical tests}
\begin{figure}[tbp]
\centering
\includegraphics[width=\linewidth]{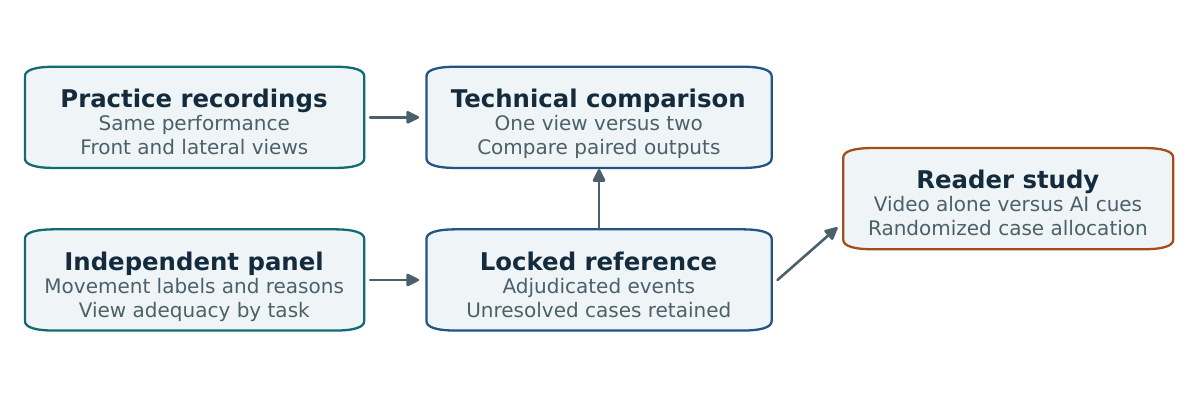}
\caption{Proposed validation data flow. The independent panel supplies the common event reference and per-view adequacy labels. The technical comparison pairs camera configurations on the same performance. The reader comparison allocates cases across conditions without repeat exposure for a given reader. The locked reference is used only to assess their responses.}
\label{fig:protocol}
\end{figure}

\subsection{Reference annotation and data splits}
Develop a form manifest with qualified judges and coaches. Define observable labels, occurrence anchors, allowed variants, reasons for an uncertain decision, and clip provenance. Figure~\ref{fig:protocol} shows how the panel records support the technical comparison and reader study. Collect consented adult practice routines with fixed frontal and lateral cameras, both showing the feet and the full body. Verify time offsets and residual synchronization errors before comparing measurements across views. Preserve raw timestamps in seconds and report onset and offset errors in milliseconds. Collect ordinary routines for the principal evaluation and intentionally difficult cases for a separate stress test. Enriched omissions, occlusions, and repetitions in the stress test do not estimate their competition prevalence.

Two or more qualified annotators independently mark movement identity, start and end, event type, visibility, and rationale before seeing model output. Retain the initial labels. An adjudicator or panel may resolve disagreements with documented reasons, with a separate uncertain disposition for unresolved cases. Record when adjudication uses a better view than the system. If adjudicators have no evidence for an underlying event, report the item as unresolved and retain it for evaluating the system's handling of uncertainty. Use a physically observable, independently reviewed trace as the movement reference where one is available. A judge's unexplained total score is unsuitable as an occurrence-level reference.

Group all views and attempts from an athlete in one partition. Keep development, tolerance selection and held-out evaluation disjoint. Freeze the pose estimator, candidate segmentation, versioned examples and thresholds before the held-out evaluation. Where practical, hold out an entire venue and report venue-specific results, including deviations in camera angle, lighting, uniform and visibility. Record participant and event counts by split and report sparse cases explicitly. The public Taegeuk dataset has a versioned repository record \citep{zhengDataset}. It can support exploratory work within the scope described by the data authors. It should be treated as development data if it informs choices.

\subsection{Unit of annotation and review record}
Record an occurrence identifier even when it is normal or insufficiently observed. For possible extras, define opportunities using prespecified intervals between reference anchors. Annotators may use all authorized views to establish the best available reference event label. They then assess whether the specific test view supports that label on its own. This yields a distinction between event validity and evidence sufficiency. If independent annotators disagree about timing, preserve the individual intervals as well as the adjudicated interval, and state the rule for tolerance-based comparison.

\begin{table}[htbp]
\centering\small
\caption{Proposed fields in the annotation and review record.}
\label{tab:record}
\begin{tabularx}{\textwidth}{@{}p{.23\textwidth}Y Y@{}}
\toprule
Record group & Required fields & Purpose\\
\midrule
Reference & Form, edition, occurrence or interval identifier & Reconstruct the comparison unit.\\
Observation & Source clip, view, timestamps, visible landmarks and anchors & Verify what the review system could observe.\\
Disposition & Candidate label or abstention, rationale, evidence window & Check whether the finding follows from the supplied video.\\
Human review & Independent labels, adjudication status and reason, adjudicated interval & Preserve disagreement and trace a final label to evidence.\\
\bottomrule
\end{tabularx}
\end{table}

Set metrics before viewing held-out labels. Include correct movements when defining the opportunities for selective assessment. Log system output for low-visibility footage even when no event label can be established. In that stratum, report the proportion of cases receiving a definite assertion despite the expert adequacy label. This is a measure of unsupported assertion. It is distinct from event error, which requires a reference event label.

\subsection{Comparators and outcomes}
Start by comparing a single-exemplar constrained DTW implementation, a movement edit aligner that returns its minimum-cost path, and the proposed multi-exemplar selective procedure. For the primary single-best-path comparison, hold the reference examples, pose inputs and visibility rules fixed, and change only whether the decision uses the best path or the set of admissible paths. Apply the evidence gate to both conditions. Use separate gate-on and gate-off analyses to estimate the effect of the gate. Use shared detections and segmentation where they are compatible. Report end-to-end accuracy separately from an analysis using human-verified boundaries. Compare one view against two on the same performance and include an ablation of the visibility gate and the plausible-path tolerance. Ordinary DTW yields temporal correspondence. For any event-level comparison, prespecify how its path is converted to extra and omitted segment candidates, and otherwise compare its alignment outcomes.

For detection metrics, match predicted and adjudicated events one to one, using event type and a preregistered temporal tolerance. Use an occurrence ID and its anchored neighborhood for omissions. Define false alerts per routine, per-class precision and recall, and onset and offset error quantiles for matched events. Any unreported labeled event, including an abstention, is a miss for end-to-end recall. Report that recall alongside conditional error among issued assessments. For a matched-coverage baseline, vary a predefined scalar quality threshold on the single-path output, using the same visibility features and development split. Report the observed test coverage of each operating point. For a fixed, model-independent set \(\mathcal{U}\) of evaluable decision opportunities with reference labels \(y_u\), define
\begin{equation}
\widehat{\mathrm{Coverage}}=
 \frac{\sum_{u\in\mathcal U}\ind\{\widehat y_u\ne\NA\}}{|\mathcal U|},
\quad
\widehat{\mathrm{Risk}}=
 \frac{\sum_{u\in\mathcal U}\ind\{\widehat y_u\ne\NA,\widehat y_u\ne y_u\}}
      {\sum_{u\in\mathcal U}\ind\{\widehat y_u\ne\NA\}}.
\label{eq:metrics}
\end{equation}
Risk is undefined at zero coverage, and coverage is undefined if the eligible set is empty. For unresolved reference labels, report the count and distribution of system decisions separately. Define the evaluation strata and denominators before unblinding. Report risk--coverage curves over decision settings and error rates by visibility stratum. For the paired camera contrast, fix the underlying opportunity set across views. At held-out opportunity \(u\), denote issuance in the two-view and single-view conditions by \(A_u^{(2)}\) and \(A_u^{(1)}\), respectively. The difference in coverage is
\begin{equation}
\widehat{\Delta}_{\mathrm{coverage}}
=\frac{1}{|\mathcal U|}\sum_{u\in\mathcal U}
 \bigl(A_u^{(2)}-A_u^{(1)}\bigr).
\end{equation}
Report this difference with both risk estimates and counts of discordant issuance, since identical aggregate coverage can hide different coverage by case. Resample whole athletes with all their cases and view conditions intact. Use athlete-clustered paired resampling to estimate intervals when enough independent athletes are available. Summarize venue effects descriptively if there are too few venues for inference. Set sample sizes and acceptance thresholds before testing using the desired uncertainty on event-level estimates and an independent recruitment feasibility check.

\subsection{Judge reader study}
Compare ordinary video review with the same video interface plus the AI-generated clip cues and reasons for abstention. Independent adjudicators define the reference before the study. Randomize and counterbalance condition assignment across qualified readers and case sets. Each reader sees each case only once. Each case is assessed in both conditions across readers. Hide the reference from readers and blind assessment of written rationales to study condition where feasible. Preserve raw responses and exposure to any erroneous suggestions.

The primary endpoint is the proportion of all assigned reader--case decisions meeting a prespecified rubric for reference-supported observations and reasons. On cases the panel marks as insufficient, the rubric requires recognition of the missing evidence. An unsubmitted response counts as unsuccessful and is also reported separately. Secondary outcomes include time from case presentation to finalized response, rate of appropriately unresolved decisions, and acceptance of incorrect cues. Use a crossed reader and case analysis that accounts for repeated performances from the same athlete where applicable. Agreement may rise because readers share a cue, so agreement alone cannot establish correctness. Experimental work on AI decision aids in other domains motivates measuring whether reviewers accept incorrect suggestions \citep{bucinca2021}. If a follow-on study asks readers for an initial judgment before revealing the cue, analyze it as a separate interface intervention because the extra step can change review time and deliberation. An immediate improvement in case review would support a decision-aid claim. Demonstrating learning requires a separate follow-up on new clips without the aid.

\section{Discussion}
\subsection{Limitations and governance}
The proposed gate relies on visibility and segmentation definitions that may differ across staff, forms and views. A common pose estimator can fail in both cameras for similar reasons. Camera checks and abstention may reduce coverage, especially in groups or conditions less represented by the development data. Record and inspect such differences before considering a new setting. A stable alignment reflects the chosen cost and reference examples. Expert review, independent annotations and sensitivity analysis remain necessary to assess its relevance.

Before data collection, document ethics review where applicable, informed consent, recording and reuse permissions, retention limits and access roles. Keep the reference edition, model identifier, timestamps, video provenance, overlays and reviewer rationale in the trace. Provide a way to correct an erroneous cue and retain the history. The pilot serves coaching and judge education. Official scoring, athlete classification and disciplinary decisions remain outside the evaluation.

\subsection{Conclusion}
Recognized Poomsae offers a prescribed sequence that can make AI-assisted video review testable at the movement level. Explicit edit operations expose possible gaps, and alternative alignment paths show when a specific attribution depends on the selected path. The evidence test prevents an apparently stable alignment from standing in for a usable view. The next empirical question is whether this combination yields accurate movement-level observations from video and improves review under controlled testing with coaches and judges.

\section*{Data and code availability}
No participant recordings were collected or analyzed for this paper. The worked example and tolerance plots use abstract symbols and specified edit costs. The workflow and validation figures diagram the proposed method and study design. The data article and repository record describe the public dataset. Researchers should confirm allowable reuse with the data provider before using the recordings. The accompanying source package includes a deterministic check of the worked example, the edge-support calculation, and the threshold in Eq.~\eqref{eq:toy-threshold}, as well as the source for all figures. 
\section*{Acknowledgments}
Thanks to GM Ramin Ray, Sport Director of Austrian Taekwondo Federation for his advice and corporations.

\clearpage\begingroup\footnotesize\raggedright\setlength{\bibsep}{2.5pt}\bibliographystyle{unsrtnat}
\bibliography{references}\endgroup

@misc{wtRules,
 author = {{World Taekwondo}},
 title = {Poomsae Competition Rules and Interpretation},
 year = {2024},
 note = {In force as of 30 September 2024. \href{https://www.worldtaekwondo.org/att_file/documents/Poomsae\%20Competition\%20Rules\%20and\%20Interpretation\%20\%28In\%20force\%20as\%20of\%20September\%2030\%202024\%29.pdf}{Rulebook PDF}}
}

@misc{wtHandbook,
 author = {{World Taekwondo}},
 title = {World Taekwondo Recognized Poomsae Handbook},
 year = {2026},
 note = {Version 1.0, July 2026. Copy hosted by Taekwondo Bond Nederland. \href{https://www.taekwondobond.nl/wp-content/uploads/World-Taekwondo-Recognized-Poomsae-Handbook-version-1.0-July-2026.pdf}{Handbook PDF}}
}

@misc{wtPara,
 author = {{World Taekwondo}},
 title = {World {Para} Taekwondo Poomsae Competition Rules},
 year = {2024},
 note = {In force as of 30 September 2024. \href{https://www.worldtaekwondo.org/att_file/documents/World\%20Para\%20Taekwondo\%20Poomsae\%20Competition\%20Rules\%2020240930.pdf}{Rulebook PDF}}
}

@article{hoang2023,
 author = {Hoang, Thi Thuy and Ahn, Heejune},
 title = {An Evaluation Method of Taekwondo Poomsae Performance},
 journal = {Journal of Information and Communication Convergence Engineering},
 year = {2023},
 volume = {21},
 number = {4},
 pages = {337--345},
 doi = {10.56977/jicce.2023.21.4.337}
}

@article{lee2020,
 author = {Lee, Jinkue and Jung, Hoeryong},
 title = {{TUHAD}: Taekwondo Unit Technique Human Action Dataset with Key Frame-Based {CNN} Action Recognition},
 journal = {Sensors},
 volume = {20},
 number = {17},
 pages = {4871},
 year = {2020},
 doi = {10.3390/s20174871}
}

@article{huang2026,
 author = {Huang, Chien-Chung and Zheng, Qing-Wei},
 title = {A dual-view videos and two-dimensional pose annotation dataset of Taekwondo Poomsae ({Taegeuk} 1--8) for human motion analysis},
 journal = {Data in Brief},
 year = {2026},
 volume = {69},
 pages = {113324},
 doi = {10.1016/j.dib.2026.113324}
}

@incollection{arriola2025,
 author = {Arriola-Pezt{\~n}a, Heriberto and Gonz{\'a}lez-Diaz, Jorge-Ernesto and Jim{\'e}nez-Nieto, Yara-Anah{\'i} and Rodr{\'i}guez-Parada, Adolfo and Mancilla-G{\'o}mez, Martin and Rodr{\'i}guez-V{\'a}squez, Gabriel},
 title = {Computer Vision Applied to Sports: A Case Study for Evaluating the Accuracy of Poomsae {Pyongwon}},
 booktitle = {New Challenges in Software Engineering},
 editor = {Mej{\'i}a, Jezreel and Mu{\~n}oz, Mirna and Rocha, Alvaro and Matos, Fernando and Espinosa-Faller, Francisco Javier and Trejo-Sanchez, Joel Antonio},
 series = {Studies in Computational Intelligence},
 volume = {1209},
 publisher = {Springer},
 address = {Cham},
 pages = {469--485},
 year = {2025},
 doi = {10.1007/978-3-031-90310-6_29}
}

@article{hua2026,
 author = {Hua, Junyi and Chen, Yiwen and Tang, Chaosheng and Ye, Jiani and Zhu, Dinghuang and Yu, Shengming},
 title = {A Systematic Review of Validation Evidence for Movement Sensing and Rule-Based Scoring in Taekwondo Poomsae},
 journal = {Sensors},
 volume = {26},
 number = {23},
 pages = {7420},
 year = {2026},
 doi = {10.3390/s26237420}
}

@article{sakoe1978,
 author = {Sakoe, Hiroaki and Chiba, Seibi},
 title = {Dynamic programming algorithm optimization for spoken word recognition},
 journal = {IEEE Transactions on Acoustics, Speech, and Signal Processing},
 year = {1978},
 volume = {26},
 number = {1},
 pages = {43--49},
 doi = {10.1109/TASSP.1978.1163055}
}

@article{needleman1970,
 author = {Needleman, Saul B. and Wunsch, Christian D.},
 title = {A general method applicable to the search for similarities in the amino acid sequence of two proteins},
 journal = {Journal of Molecular Biology},
 year = {1970},
 volume = {48},
 number = {3},
 pages = {443--453},
 doi = {10.1016/0022-2836(70)90057-4}
}

@inproceedings{geifman2017,
 author = {Geifman, Yonatan and El-Yaniv, Ran},
 title = {Selective Classification for Deep Neural Networks},
 booktitle = {Advances in Neural Information Processing Systems},
 volume = {30},
 year = {2017},
 note = {\href{https://papers.neurips.cc/paper_files/paper/2017/hash/4a8423d5e91fda00bb7e46540e2b0cf1-Abstract.html}{Proceedings}}
}

@article{jeon2026,
 author = {Jeon, Min-Woo and Kim, Hong-Suk and Park, Ji-Yong and Cho, Hye-Soo},
 title = {Revisiting judging reliability in taekwondo freestyle Poomsae: implications for {AI}-supported evaluation},
 journal = {Frontiers in Psychology},
 year = {2026},
 volume = {17},
 pages = {1881296},
 doi = {10.3389/fpsyg.2026.1881296}
}

@misc{zhengDataset,
 author = {Zheng, QingWei},
 title = {Taekwondo Poomsae (8 Forms): Front and Side Videos with 12-Keypoint Pose Annotations ({MediaPipe})},
 year = {2025},
 note = {Mendeley Data, version 1. Dataset DOI: \href{https://doi.org/10.17632/bjy7vr4xkt.1}{10.17632/bjy7vr4xkt.1}}
}

@inproceedings{xu2022,
 author = {Xu, Jinglin and Rao, Yongming and Yu, Xumin and Chen, Guangyi and Zhou, Jie and Lu, Jiwen},
 title = {{FineDiving}: A Fine-Grained Dataset for Procedure-Aware Action Quality Assessment},
 booktitle = {Proceedings of the IEEE/CVF Conference on Computer Vision and Pattern Recognition},
 pages = {2949--2958},
 year = {2022},
 note = {\href{https://openaccess.thecvf.com/content/CVPR2022/html/Xu_FineDiving_A_Fine-Grained_Dataset_for_Procedure-Aware_Action_Quality_Assessment_CVPR_2022_paper.html}{CVF proceedings}}
}

@inproceedings{abufarha2019,
 author = {Abu Farha, Yazan and Gall, J{\"u}rgen},
 title = {{MS-TCN}: Multi-Stage Temporal Convolutional Network for Action Segmentation},
 booktitle = {Proceedings of the IEEE/CVF Conference on Computer Vision and Pattern Recognition},
 pages = {3575--3584},
 year = {2019},
 note = {\href{https://openaccess.thecvf.com/content_CVPR_2019/html/Abu_Farha_MS-TCN_Multi-Stage_Temporal_Convolutional_Network_for_Action_Segmentation_CVPR_2019_paper.html}{CVF proceedings}}
}

@article{kim2026,
 author = {Kim, Hyeong-Ju and Park, Ju-Sik},
 title = {Analysis of Scoring Differences by Scoring Item According to the Competition Stage of Taekwondo's Official Poomsae},
 journal = {Journal of Coaching Development},
 volume = {28},
 number = {3},
 pages = {224--231},
 year = {2026},
 note = {\href{https://www.kci.go.kr/kciportal/ci/sereArticleSearch/ciSereArtiView.kci?sereArticleSearchBean.artiId=ART003320243}{Korean Citation Index record}}
}

@article{bucinca2021,
 author = {Bu{\c{c}}inca, Zana and Malaya, Maja Barbara and Gajos, Krzysztof Z.},
 title = {To Trust or to Think: Cognitive Forcing Functions Can Reduce Overreliance on {AI} in {AI}-Assisted Decision-Making},
 journal = {Proceedings of the ACM on Human-Computer Interaction},
 volume = {5},
 number = {CSCW1},
 year = {2021},
 doi = {10.1145/3449287}
}

@article{choi2025,
 author = {Choi, Taeseok and Heo, Seoyoon},
 title = {Towards Automated Taekwondo Poomsae Scoring: Evaluating Modern Neural Network Techniques},
 journal = {Journal of International Academy of Physical Therapy Research},
 volume = {16},
 number = {1},
 pages = {3401--3409},
 year = {2025},
 doi = {10.20540/JIAPTR.2025.16.1.3401}
}

@incollection{fukushima2024,
 author = {Fukushima, Takashi and Haggenmueller, Klaus and Lames, Martin},
 title = {Validity of {OpenPose} Key Point Recognition and Performance Analysis in Taekwondo},
 booktitle = {Proceedings of the 14th International Symposium on Computer Science in Sport},
 series = {Lecture Notes in Data Engineering and Communication Technologies},
 volume = {209},
 pages = {68--76},
 publisher = {Springer Nature Singapore},
 year = {2024},
 doi = {10.1007/978-981-97-2898-5_8}
}
\end{document}